\documentclass[IEEE Transactions on Instrumentation and Measurement]{IEEEtran}

\usepackage[bookmarks=false]{hyperref}
\usepackage{graphicx}
\usepackage{amssymb}
\usepackage{latexsym}
\usepackage{floatrow}
\usepackage{url}
\usepackage{xcolor}
\usepackage{hyperref}
\usepackage{framed,multirow}
\usepackage[ruled,vlined]{algorithm2e}
\usepackage{booktabs}
\usepackage{multicol}
\usepackage{lipsum}
\usepackage{mwe}
\usepackage{multirow}
\usepackage{relsize}
\usepackage{float}
\usepackage{floatrow}
\usepackage{verbatim}
\usepackage{tabularx}
\usepackage{amsmath,epsfig}
\usepackage{bbm}
\usepackage{relsize}
\usepackage{booktabs}
\usepackage{amssymb}
\usepackage{latexsym}
\usepackage{upgreek}
\usepackage{cite}
\usepackage{subcaption}
\usepackage{xcolor}
\ifCLASSINFOpdf
\else
\fi
\begin{document}
%
\title{MSAD-Net: Multiscale and Spatial Attention-based Dense Network for Lung Cancer Classification}
%
%
%

\author{Santanu Roy, Shweta Singh, Palak Sahu, Ashvath Suresh, and Debashish Das 
\thanks{Dr. Santanu Roy, a former employer of Christ (Deemed to be University), is now working at the Computer Science and Engineering Department, NIIT University, Jaipur, India (Email: santanuroy35@gmail.com). 

Shweta Singh, and Palak Sahu are working with the Computer Science and Engineering Department, NIIT University, Jaipur, India.
(Email: Shweta.singh21@st.niituniversity.in; palak.sahu20@st.niituniversity.in). 

Ashvath Suresh is working with the CSE department, Christ (Deemed to be University), Bangalore, India. 

(Email: ashvath.suresh@btech.christuniversity.in); 

Whereas, Dr. Debashish Das is working with Faculty of Computing, Engineering and the Built Environment, Birmingham City University, UK. 
(Email:
Debashish.Das@bcu.ac.uk)
}
\thanks{Manuscript received xxxx; revised xxxx.}}

%
%

\markboth{IEEE Transactions on xxxx xxxx,~Vol.~xx, No.~x, Month~xxxx}%
{Shell \MakeLowercase{\textit{et al.}}: Bare Demo of IEEEtran.cls for IEEE Journal}
%



\maketitle

\begin{abstract}
Lung cancer, a severe form of malignant tumor that originates in the tissues of the lungs, can be fatal if not detected in its early stages. It ranks among the top causes of cancer-related mortality worldwide. Detecting lung cancer manually using chest X-Ray image or Computational Tomography (CT) scans image poses significant challenges for radiologists. Hence, there is a need for automatic diagnosis system of lung cancers from radiology images. With the recent emergence of deep learning, particularly through Convolutional Neural Networks (CNNs), the automated detection of lung cancer has become a much simpler task. Nevertheless, numerous researchers have addressed that the performance of conventional CNNs may be hindered due to class imbalance issue, which is prevalent in medical images. In this research work, we have proposed a novel CNN architecture ``Multi-Scale Dense Network (MSD-Net)'' (trained-from-scratch). The novelties we bring in the proposed model are (I) We introduce novel dense modules in the $4^{th}$ block and $5^{th}$ block of the CNN model. We have leveraged 3 depthwise separable convolutional (DWSC) layers, and one 1$\times$1 convolutional layer in each dense module, in order to reduce complexity of the model considerably. (II) Additionally, we have incorporated one skip connection from $3^{rd}$ block to $5^{th}$ block and one parallel branch connection from $4^{th}$ block to Global Average Pooling (GAP) layer. We have utilized dilated convolutional layer (with dilation rate=2) in the last parallel branch in order to extract multi-scale features. Extensive experiments reveal that our proposed model has outperformed latest CNN model ConvNext-Tiny, recent trend Vision Transformer (ViT), Pooling-based ViT (PiT), and other existing models by significant margins. All the codes of these experiments along with its graphs and confusion matrix are provided on a GitHub link:

\textbf{https://github.com/Singhsshweta/Lung-Cancer-Detection}.
\end{abstract}


%
\IEEEpeerreviewmaketitle

\section{Introduction}
\IEEEPARstart{L}ung cancer is a dangerous form of malignant neoplasm because it originates in the lung tissues [1], which are vital organs responsible for supplying oxygen to the body. The term ``malignant'' means that the cancerous cells are aggressive and capable of spreading (or, metastasizing) to other parts of the body [2]. If lung cancer is not detected at early stage, the tumor can grow and invade nearby tissues or spread to distant organs through the bloodstream (or, lymphatic system), making treatment more difficult. According to WHO, lung cancer was the leading cause of cancer-related deaths worldwide in 2020 [3]. Computed Tomography (CT) is one of the most reliable techniques [4] for the early detection of lung cancer. It is particularly effective in identifying small nodules or tumors in the lungs that may not be visible on standard chest X-rays. However, detecting lung cancer in its early stages using CT scans presents substantial challenges for radiologists, as early tumors can be typically small, inconspicuous, and may resemble normal anatomical features. Consequently, there is a growing need for Computer-Aided Diagnosis (CAD) system [5] to accurately detect lung cancers from CT scan images. The emergence of deep learning and CNN models, has revolutionized the automatic diagnostic capabilities, especially,  in the field of medical imaging domain. The potential of such advancements in medical diagnosis is profound. Moreover, the ability of automatically extracting features of CNN models enables the CAD system to enhance the efficacy of identifying suspicious lesions and furthermore, can accurately classify different types of lung cancers. In this study, we examine various forms of lung cancer, including Adenocarcinoma, Large Cell Carcinoma, and Squamous Cell Carcinoma [6]. Each of these tumor types requires different treatment strategies, depending on the stage and location of the tumor, as well as the patient's overall health condition. Therefore, early detection of these lung cancers is crucial for physicians to determine further treatment options and prognosis.

Afshar et al. [7] (2021) introduced an innovative MIXCAPS architecture that integrates capsule networks with an expert model to efficiently predict malignancy from CT scans. Capsule network [8] is an advanced deep learning framework aimed at addressing certain limitations of conventional CNNs, by incorporating a group of neurons that facilitate extracting hierarchical and spatial features (of lung nodules), thereby significantly improving the efficacy of the model. Ashnil Kumar et al. [9] (2019) proposed a multimodal CNN framework that extracts features from both CT and PET images through two parallel encoders. These features are subsequently combined using a fusion module, followed by a reconstruction module. By fusing features from both modalities, the CNN model will be able to extract a diverse set of features compared to handling each modality separately, thereby it enhances the generalization ability of the model. N. Faruqui et al. [10] (2021) introduced a hybrid deep CNN model ``Lung-Net" that integrates CNN-derived features with valuable features from medical IoT (MIoT) data based on wearable sensors. With extra physician-assisted MIoT data, their model achieved a performance boost of 2-3\%. Md.S. Hossain et al. [11] (2025) proposed an ensemble methodology in which the final classification decision is determined by majority voting between two models: (I) a Convolutional Neural Network (CNN), inspired by VGG-16 architecture, and (II) a Support Vector Machine (SVM). Additionally, they utilized Contrast Limited Adaptive Histogram Equalization (CLAHE) and Singular Value Decomposition (SVD) as a pre-processing step. P.M. Shakeel et al. [12] have introduced a novel pipeline (of series of image processing) for automatic diagnosis of lung cancer from Cancer imaging Archive (CIA) dataset. First, in order to reduce noise they deployed a weighted mean histogram equalization, moreover, they have applied profuse clustering technique (IPCT) for segmenting the affected region from lung modules and then classify segmented images using a conventional CNN model. All the aforementioned approaches introduced innovations either through designing novel CNN model architectures, or by employing an extensive range of image processing techniques as a pre-processing step to mitigate the challenges of dataset. The former approach appears to be a valid strategy; however, the latter heavily relies on image processing methods ([11],[12]) making it not always feasible. This is because a specific pipeline of such techniques may perform effectively on a particular dataset, but there is no assurance that the same will generalize across diverse CT image datasets. Many more studies on lung cancer detection using CNN models can be found in the literature review from [13]-[17].

Another effective approach is to utilize cutting-edge technology ``Vision Transformer (ViT) model" which is inspired from the concept of self-supervised model [18]. Arvind Kumar et al. [19] (2024) recently employed a ViT model on histopathology images, to classify several grades of lung cancer. Their proposed model is very similar to the conventional ViT model [20], in which patch size 16$\times$16 is utilized along with positional embedding. The advantage of using this ViT model is that unlike CNN model, it can automatically provide more attention to the specific fine-grained features (in small lung modules) which contain crucial information for lung cancer classification. However, the conventional ViT is computationally costly, thus, often prone to overfitting for small medical datasets. Another limitation of this ViT is that, like CNN it does not have multiscale hierarchical structure [21], which is advantageous for the object classification task. Therefore, numerous researchers [22-23] have proposed hybrid kinds of transformer models that integrates both the principles of CNN and ViT. M. Imran et al. [23] (2024) introduced a hybrid CNN-ViT architecture for classifying three types of lung cancer from histopathology images. Their approach employed multi-scale patches (of varying dimensions) as input to the ViT module, resulting in an effective multi-scale ViT framework. In addition to this, they incorporated CNN layers parallel to this multi-scale ViTs, to extract both local and global features from lung modules. Although their model seems like a valid approach, the extensive parallel concatenation of ViT encoders and CNN modules renders the overall architecture intricate and computationally expensive. B.Heo et al. [24] (2022) proposed a pooling-based ViT (named ``PiT"), that incorporates pooling layers interchangeably into the ViT model. As a consequence, it has now the multiscale hierarchical structure like CNN, moreover, computational complexity in the self-supervised encoder has been considerably reduced due to dimensional reduction by pooling layers.

Other prominent attention mechanisms designed for image classification are the Squeeze-and-Excitation Network (SE-Net) [25] and the Convolutional Block Attention Module (CBAM) [26], which are incorporated into CNN model. J. Hu et al. [25] first time proposed an SE-Net module to capture channel-wise inter-dependencies, thereby enhancing the overall performance of the CNN model. 
Z. Xu et al. [27] (2022) proposed a novel CNN architecture influenced by Inception-V3, termed ``ISA-Net," which was trained from scratch. ISA-Net employs only two Inception unit blocks alongside standard convolutional layers. Moreover, a Channel Attention (CA) module (influenced by SE-Net) and a Spatial Attention Module (SAM) are integrated in parallel and utilized within the deeper layers of their proposed model. Likewise, H. Xiao et al. [28] (2023) proposed a multi-feature multi-attention network (MFMA Net) to classify several types of lung tumors efficiently from CT images. Their model architecture includes a multiscale spatial-channel attention module that substitutes GAP with scale-varied spatial-pooling techniques. This scale-varied pooling technique enables the model to extract multiscale features in both channel and spatial dimension. S.Roy et al. [29] (2024) first time observed that integrating such CBAM or SE-Net modules into deeper layers of a CNN model could lead to substantial information loss. To address this, they proposed a novel spatial attention module (SAM) in which they utilize 3$\times$3 dilated convolutional filters (with dilation rate 2) instead of standard 7$\times$7 convolutional filter. These dilated convolutional filters with padding "same" preserves information in spatial dimension, as well as it reduces the computational complexity of the SAM attention block. Our proposed attention module is inspired by the aforementioned work; however, its overall architecture is different from SAM-Net [29]. More attention-based frameworks, proposed on radiology images, can be found in [30-32]. 
\begin{figure*}[h]
		\centering
		\includegraphics[width=17.4cm,height=7.4cm]{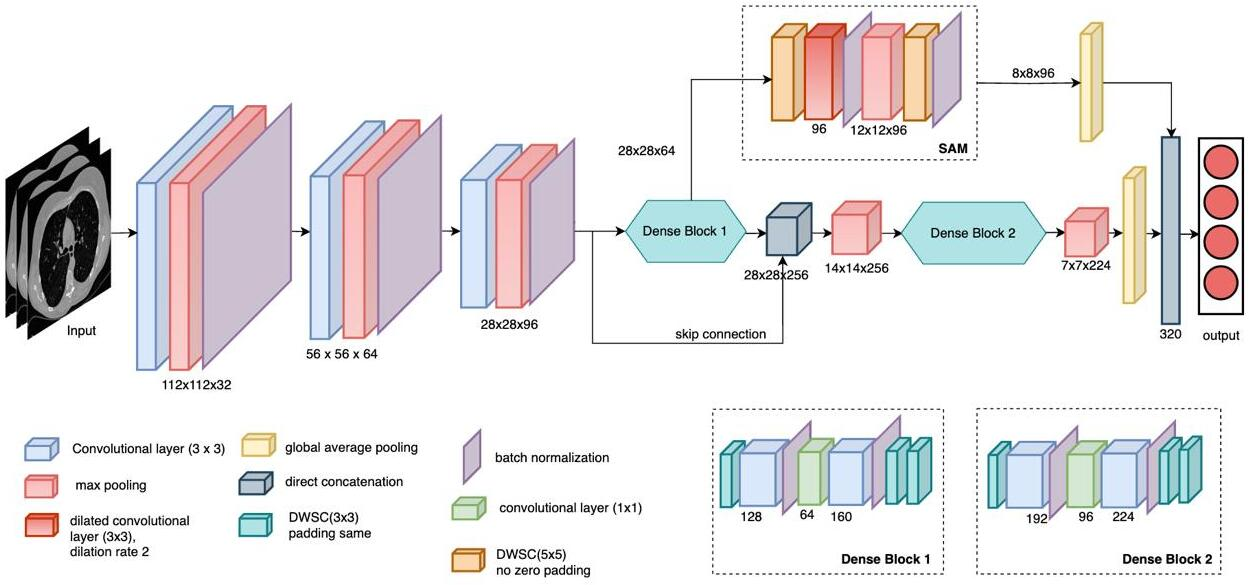}
		\caption{Entire Block Diagram of the proposed MSAD-Net \textbf{(zooming is preferable)}}
	\end{figure*}
    
\vspace{0.1cm} 
\textbf{Contributions of the paper}
The contributions of this paper are explained as follows: 
\begin{enumerate}
        \item A novel CNN architecture ``Multi-Scale and Spatial Attention Dense Network (MSAD-Net)'' is proposed. MSAD-Net consists of novel dense modules in which DWSC layers and 1$\times$1 layer are utilized to significantly reduce the number of trainable parameters. 

	\item A spatial attention module (SAM) is also incorporated in parallel to the proposed base model, where a 3$\times$3 dilated convolutional layer is applied with a dilation rate of 2. This enables the parallel branch to capture multi-scale features with reduced computational cost, thereby enhancing the efficacy of the model. 

 \item Extensive experiments reveal that the proposed ``MSAD-Net'' outperforms recent trend ``Vision Transformer (ViT)", ``Pooling-based ViT (PiT)", and the latest CNN model ``ConvNext-Tiny", by significant margins. 
 
 \item The proposed model ``MSAD-Net'' was tested on two diverse CT datasets for checking the validity. Additionally, a 5-fold cross-validation experiment was conducted, and explainable AI is implemented in order to prove the model's validity.
\end{enumerate}

The rest of the paper is organized in the following way: Section-II explains the entire proposed methodology along with parameter calculations in dense module. In Section-III, the result of the proposed framework is compared with the results of the benchmark methods. In Section-IV, we present our concluding remarks and future work directions. 

\section{Proposed Methodology}
The proposed methodology part can be further divided into three parts: (a) Base Model Architecture, (b) Mathematical Analysis for parameter calculations in dense module, (c) Spatial Attention Module.  

\subsection{Base Model Architecture:}
The proposed ``MSAD-Net'' architecture is inspired from the residual network [33], a parallel concatenation CNN [34], and SAM-Net [29]. The proposed base model consists of only 5 convolutional blocks which is followed by Global Average Pooling layer and Softmax layer. In convolutional blocks 1 to 3, we used an identical structure, consisting of a single convolutional layer with 3×3 kernel size with padding `same', followed by a batch normalization layer and one max-pooling layer. The number of filters in block 1 to block 3 for convolutional layer is chosen 32, 64 and 96 respectively. Batch normalization layer converts the scattered 2D tensor input (after convolution) into a normalized distribution having mean $0$ and standard deviation $1$. It ensures a smooth gradient flow throughout the network and hence, reduces the over-fitting problem a bit.

In the $4^{th}$ and $5^{th}$ block of the proposed model, we introduce novel dense modules. Each dense module comprises one 3$\times$3 depth-wise separable convolutional (DWSC) layer [35], followed by two 3$\times$3 convolutional layers, with a 1$\times$1 layer sandwiched between them. Eventually, the dense module ends with two additional DWSC layers. This dense block, as shown in Fig. 1, is illustrated separately on the bottom and right-hand sides of the same figure. The purpose of inserting the 1$\times$1 layer in between these two 3$\times$3 convolutional layers is that it will decrease the spectral dimension of convoluted output, thereby significantly reducing the overall number of trainable parameters. A mathematical analysis is presented to substantiate this claim in the next subsection. Across all the 3$\times$3 and 1$\times$1 convolution layers, as well as the DWSC layers (in the dense module), the ReLU activation function is deployed, additionally, padding ``same'' is utilized across all these layers. 
\begin{figure*}[h]
		\centering
		\includegraphics[width=17.4cm,height=7 cm]{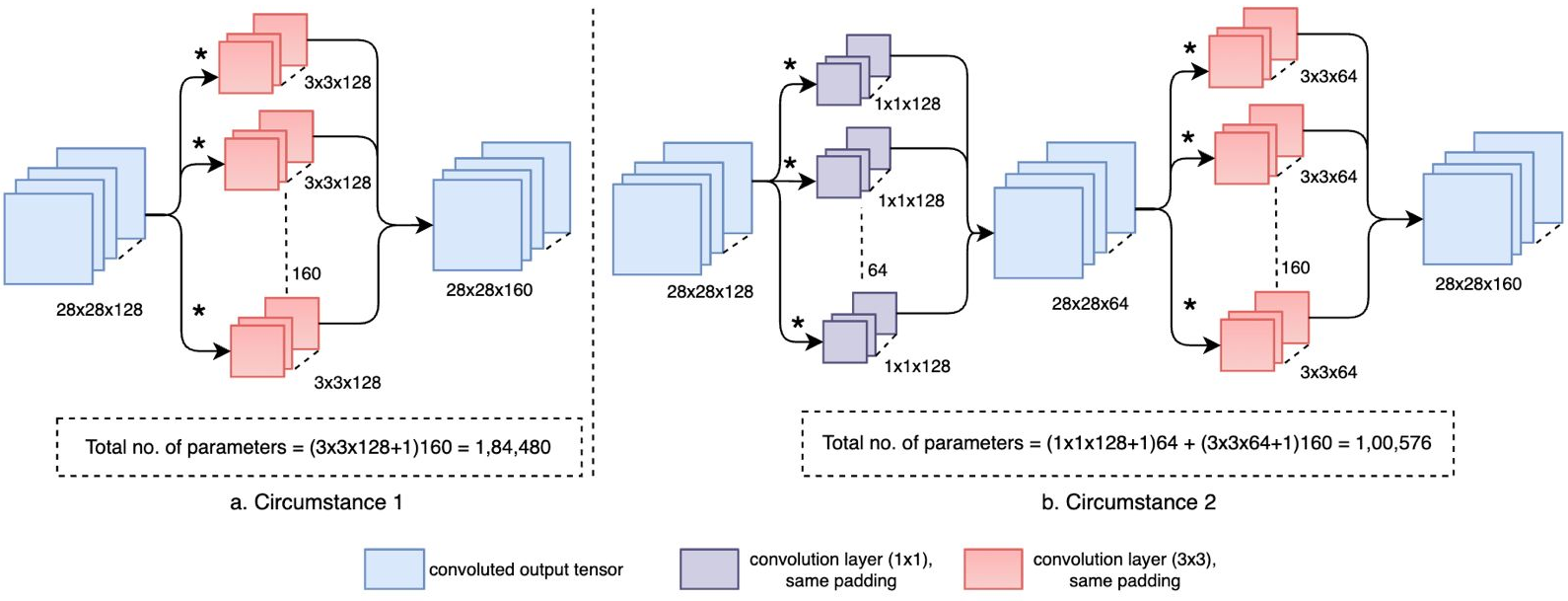}
		\caption{Entire Split of Convolutional layers inside Dense block 1: (a) Circumstance 1 reflects if we employ two conv layers back to back in Dense Block 1, (b) Circumstance 2 reflects when 1$\times$1 conv layer is inserted in between two conv layers in Dense Block 1}
	\end{figure*}
The max-pooling layer is not employed between two convolutional layers inside a dense module. Instead, it is applied after the end of each dense module. From design perspective, this offers certain advantages, such as enabling the flexibility to establish a direct skip connection from the input to the output of the dense block. We have employed first skip connection from the output of $3^{rd}$ block to the input of $5^{th}$ block as depicted in Fig.1. We have leveraged this skip connection because it enables the neural network, improving the overall gradient flow [33] and generalization ability of the model to some extent.
    
\subsection{Parameter calculations in the Dense Module}
From Fig.1 it is evident that we have employed back to back two 3$\times$3 convolutional layers, with a 1$\times$1 layer placed between them in the dense module. The convoluted tensor output after 3$\times$3 convolutional layer, can be represented by 

\begin{equation}
		O_{c}(s)={ReLU({\sum_{i=1}^{f_1}{{D_i(\tau)}_{3\times3}}*{I(s)}_{h\times h}}+b)}  
\end{equation}
Here, in equation (1), `$*$' indicates convolution operation, $I(s)$ is the original image having size $h\times h$, ${D_i(\tau)}$ is the filter co-efficent of 3$\times$3 kernel, $b$ is the bias.

The number of parameters $n_c$ in this 3$\times$3 convolutional layer can be computed by the following equation, 
\begin{equation}
		n_c= (3^{2}.f_0+1).f_1
\end{equation}

Here, $f_0$ is the number of filters in the previous layer (input), $f_1$ is the number of filters in the current layer. 

On the other hand, the convoluted output tensor after the 1$\times$1 layer [35] can be represented by

\begin{equation}
		O_{1\times1}(s)={ReLU({\sum_{i=1}^{f_{1\times1}}{c_{i}.{I(s)}_{h\times h}}+b)}}  
\end{equation}

Here, in equation (3), $c_i$ is the real constant which is a trainable parameter for the $i^{th}$ filter. Here, convolution between the original image $I(s)$ and the co-efficient of 1$\times$1 filter becomes dot product because kernel size is 1$\times$1, here $b$ is the bias.

Consequently, the number of parameters $n_{1\times1}$ in the 1$\times$1 layer is considerably less and shown in equation (4). It is evident from equation (4), that $n_{1\times1}$ is considerably lesser than $n_c$.
\begin{equation}
		n_{1\times1}= (1\times1\times f_{0}+1).{f_{1\times1}}
\end{equation}
Here, $f_{1\times1}$ is the number of filters, chosen in 1$\times$1 layer, which becomes the input spectral dimension for the next 3$\times$3 convolutional layer. In other words, the number of parameters are not only less in equation (4), but also this number of filters $f_{1\times1}$ can be adjusted such a way that it also impacts heavily on the number of parameters in the next convolutional layer. This statement can be further mathematically proved under certain circumstances.

Suppose, we consider the first dense block, here the input tensor after the first convolutional layer in the first dense module (in $4^{th}$ block) is $h\times h$ and spectral dimension is $128$, as depicted in Fig.2. Now, if we directly convolute another layer of $160$ filters, then the number of parameters according to the equation (2) will not be dependent on spatial dimension $h\times h$.

\begin{equation}
		n_c= (3\times 3 \times 128+1).160=1,84,480.
\end{equation}

On the other hand, if we deploy first $64$ numbers of 1$\times$1 filters and then employ $160$ numbers of convolutional filters (with 3$\times$3 kernel size) as demonstrated in Fig.1, then the number of parameters will be equal to 

\begin{equation}
		{n_c}^{'}= (3^{2}.64+1).160+(1\times 1\times 128+1)*64=1,00,576.
\end{equation}

 Comparing equation (5) and (6), it is quite evident that the total number of parameters in the $1^{st}$ dense module (or, $4^{th}$ block) is reduced by $83,904$ due to incorporating 1$\times$1 layer in between these two convolutional layers. This is more clearly depicted in the Fig.2. This difference between $n_c$ and ${n_c}^{'}$ will be more prominent in the deeper layer, that is, in block 5. 
    \begin{figure*}[h]
		\centering
		\includegraphics[width=18.2cm,height=6.3 cm]{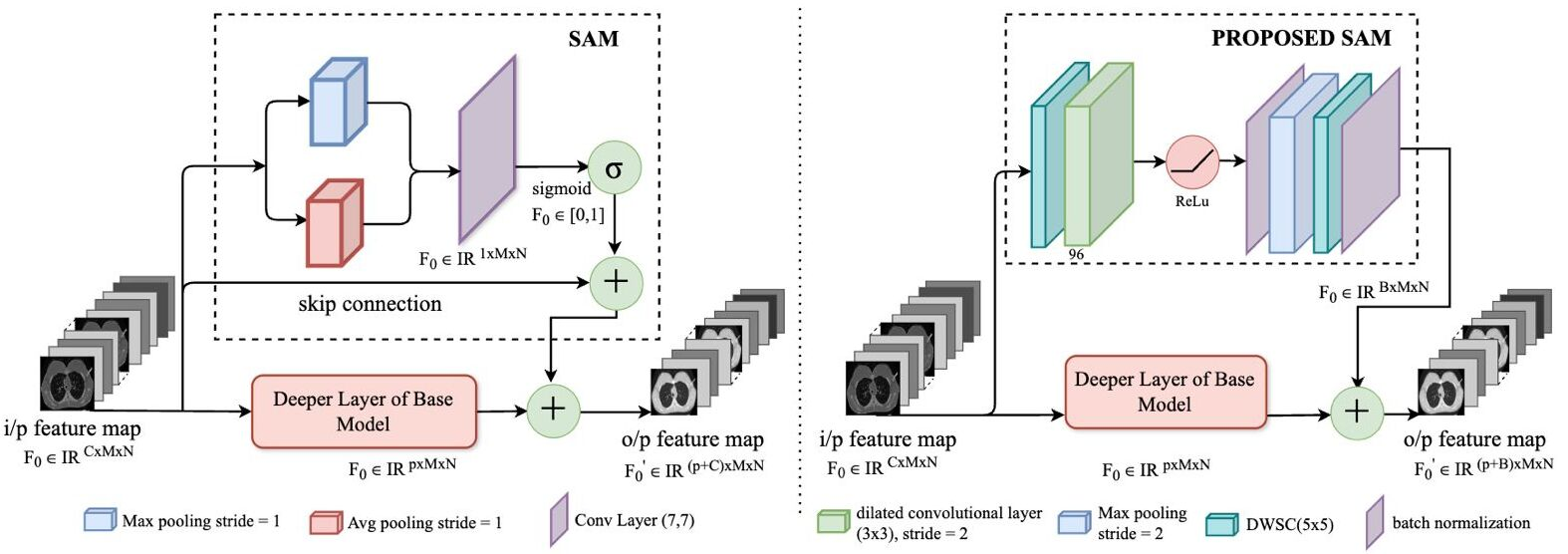}
		\caption{ Left Hand Side (LHS) image represents the conventional spatial attention module (SAM), connected parallel to the base model. RHS image represents the proposed SAM where dilated convolutional layer (3$\times$3), and DWSC (5$\times$5) are incorporated.}
	\end{figure*}
The number of Max-pooling layers employed in the base model is 5, as illustrated in the Fig.1. Given that the input tensor size is $224\times 224$, the spatial dimensions of the output from the final block will be $224/2^{5}$ x $224/2^{5}= 7\times7$. This output is then passed through a Global Average Pooling (GAP) layer [36], instead of the flatten layer. The GAP layer computes the average across the spatial dimensions, thereby reducing the number of neurons to $224$ only. Consequently, the number of neurons at the input of FC layer, is reduced (approximately) by a factor of $7\times 7=49$. 

\subsection{Proposed Spatial Attention Module (SAM)}
Another novelty of our work is that we have incorporated a $2^{nd}$ skip connection, from the middle of $4^{th}$ convolutional layer to the GAP layer as depicted in Fig.1. We call this $2^{nd}$ skip connection a spatial attention module (SAM), since it gives some attention to the global features (at multiscale) in the spatial domain. In contrast to traditional SAM (inspired from CBAM [26]), that mostly diminishes other channels to single channel, our proposed attention block exploits multiple channels. The intermediate feature map by conventional SAM is converted from $F_m\in R^{C\times M \times N}$ to $F_m\in R^{1\times M \times N}$, depicted in Fig.3a. Thus, instead of focusing on multiple channels, it now focuses only on one channel, especially giving attention in the spatial domain. On the other hand, in our proposed SAM block, feature maps are converted from $F_m\in R^{C\times M \times N}$ to $F_m\in R^{B\times M\times N}$ where $B>1$, illustrated in Fig.3b. 
\begin{equation}
  SAM: \hspace{0.1cm} R^{C\times M \times N} \rightarrow R^{B\times M \times N}
\end{equation}
We have observed that the conventional SAM emphasizes only on one channel, this often leads to substantial information loss due to avoiding other channels. To address this, the proposed SAM incorporates multiple number of channels, thereby minimizing the data loss. Moreover, the impact of this conventional SAM on the final decision of the classifier would be merely $1/(224+64+1)$, or approximately $0.34\%$ on our proposed model, as it utilizes only one channel. We have observed by several experiments that to efficiently classify numerous types of (lung) tumors, it was crucial to focus on global features within CT images. Therefore, we increase this number of channels from 1 to 96 in the proposed SAM (empirically), so that more variety of such global features would be considered at the final decision of the classifier. As a consequence, the contribution of the spatial attention in the final classifier will be $96/(224+96)=0.3$, that is, 30\%, for our proposed SAM. Hence, it significantly enhances the effect of spatial attention mechanism in the proposed MSAD-Net.

\vspace{0.1cm}
The conventional SAM block starts with the concatenation of Max-pooling and Average pooling with stride 1, followed by a single 7$\times$7 convolutional filter with padding ``same'', as shown in Fig.3a. This 7$\times$7 filter enables the model to focus on multiscale features (or global features). On the other hand, the proposed SAM block starts from a 5$\times$5 DWSC layer (with no padding). Indeed, two such DWSC layers are employed, with a dilated convolutional layer (with kernel (3,3)) positioned between them, as depicted in Fig.3b. Additionally, we inserted one Max-pooling layer ($/2$) after the DilConv layer, to make the spatial dimension of the output tensor down-sampled ($/2$) to 7$\times$7, which is feasible spatial size for the GAP layer. 

The proposed SAM can be mathematically represented by the following equation. 
\begin{equation*}
    F_{SAM}= \tau_{BN}(D_{w(5\times5)}(M_p(\tau_{BN}(f_{Dil}(D_{w(5\times5)}(F_0)))))
\end{equation*}
\begin{equation}
   where \hspace{0.2cm} F_{SAM} \in R^{B\times M\times N}
\end{equation}
Here, in equation (8), the input feature maps $F_0$ are passed through a DWSC layer with 5$\times$5 kernel $D_{w(5\times5)}(\hspace{0.1cm})$, followed by DilConv layer $f_{Dil}(\hspace{0.1cm})$, and another DWSC layer $D_{w(5\times5)}(\hspace{0.1cm})$. Here, $\tau_{BN}$ represents the batch normalization layer and $M_p$ represents the Max pooling layer with stride 2. 

A Dilated Convolutional (DilConv) layer [37] with 3$\times$3 kernel and dilation rate 2 can be represented by the following equation. 
\begin{equation}
		f_{Dil}(F_0)={ReLU({\sum_{i=1}^{B}{{C_i(\tau)}_{3\times3}}|_{dr=2}*{F_0}+b})}
\end{equation}
Here, in equation (9), '$*$' indicates dilated convolution operation, with dilation rate $dr=2$, ${C_i(\tau)}_{3\times3}$ is the (dilated) filter co-efficient in 3$\times3$ kernel, $B$ is the number of filters employed in the present (dilated) convolutional layer, that is empirically chosen $B=96$ in the proposed SAM. $F_0$ is the input tensor coming to the DilConv filter, $b$ is bias. These DilConv filters with 3$\times$3 kernel and dilation rate 2 closely resemble the behavior of 5$\times$5 convolutional filters, however, substituting 5$\times$5 filters with DilConv filters substantially reduces the overall computational cost. Thus, DilConv filters enables our proposed SAM to automatically extract multiscale features, with much less computational complexity.

On the other hand, DWSC layer [35] with 5$\times$5 kernel $D_{w(5\times5)}$, can be represented by the following equation. 
\begin{equation}
	D_{w(5\times5)}(F_0)={ReLU({\sum_{i=1}^{C}{{C(\tau)}.F_i}}+b)}
\end{equation}
Here, in equation (10), $C$ is the number of input channels coming to the SAM block, and $F_0$ is the input feature. ${C(\tau)}$ is the filter co-efficient employed only once here, which is followed by 1$\times$1 filter that does point-wise multiplication. Hence, in equation (10), dot product of ${C(\tau)}$ and $F_i$ is considered. 

Hence, the number of trainable parameters $n_D$ in the DWSC layer in equation (11), is significantly less as compared to the same of conventional 2D Conv layers, given in equation (2).
\begin{equation}
		n_D= (3^{2}.1+1).{C}=10C
\end{equation}
Hence, in the proposed SAM block two such DWSC layers are deployed to reduce the computational complexity of the model, simultaneously, the use of 5$\times$5 filters facilitates the extraction of multiscale features, thereby enhancing the effect of the spatial attention mechanism to some extent.

Another notable difference observed in Fig.3 is that, unlike the conventional SAM, our proposed SAM omits the use of the sigmoid activation function within the block. Instead, we utilize a ReLU activation followed by a batch normalization layer, which serves a similar purpose. The only distinction is that a sigmoid function introduces a greater non-linearity in the output tensor. However, we wanted to retain all essential global features extracted in the SAM block as it is, hence, we opted not to use the sigmoid activation in the proposed SAM block.

This is to clarify that recently we have proposed similar kind of SAM block in [29], which also deploys DilConv layer in the SAM block for multiscale feature extraction. However, its overall architecture slightly differs from our proposed SAM block in this study. (I) One of such difference is that we have not employed additional skip connection parallel to the SAM block. This change was made purely from a design perspective, as it offers certain advantages. (II) Unlike SAM-Net, our proposed SAM blocks employed two Depthwise Separable layers (DWSC) with 5$\times$5 kernel size (as mentioned earlier), that further boosts the impacts of multi-scale feature extraction along the path of SAM block, with less computational complexity. To validate the effectiveness of the proposed model ``MSAD-Net" over SAM-Net [29], a comparative performance analysis is also presented later in Section IV.

\begin{table*}[tb]
\begin{center}
\caption{Comparisons of the the proposed ``MSAD-Net'' with existing models on testing, for two CT datasets (\textbf{Weighted Average of the metrics is displayed}). The best results are represented by bold letters for only 4-class dataset. However, bold letters were avoided for 3-class dataset, since many models had the best results.}
\label{tab:my-table}
\resizebox{0.98\columnwidth}{!}
{
\begin{tabular}{|c|c|cccl|cccc|}
\hline
\multirow{2}{*}{Model}                                                      & \multirow{2}{*}{\begin{tabular}[c]{@{}c@{}}Training\\ Method\end{tabular}} & \multicolumn{4}{c|}{4-class CT Dataset}                                                                                                          & \multicolumn{4}{c|}{3-class CT Dataset}                                                                                   \\ \cline{3-10} 
                                                                            &                                                                            & \multicolumn{1}{l|}{Accuracy}       & \multicolumn{1}{c|}{Precision}      & \multicolumn{1}{c|}{F1-score}       & secs/ep                        & \multicolumn{1}{l|}{Accuracy}       & \multicolumn{1}{c|}{Precision}      & \multicolumn{1}{c|}{F1score}        & secs/ep \\ \hline
DenseNet-121                                                                & pre-trained                                                                & \multicolumn{1}{c|}{0.841}          & \multicolumn{1}{c|}{0.847}          & \multicolumn{1}{c|}{0.841}          &    \multicolumn{1}{c|}{22}                            & \multicolumn{1}{c|}{0.955}          & \multicolumn{1}{c|}{0.955}          & \multicolumn{1}{c|}{0.955}          &    236     \\ \hline
Inception-V3                                                                & pre-trained                                                                & \multicolumn{1}{c|}{0.944}          & \multicolumn{1}{c|}{0.951}          & \multicolumn{1}{c|}{0.945}          &    
\multicolumn{1}{c|}{11}                            & \multicolumn{1}{c|}{0.955}          & \multicolumn{1}{c|}{0.963}          & \multicolumn{1}{c|}{0.955}          &     141    \\ \hline
MobileNet-V2                                                                & pre-trained                                                                & \multicolumn{1}{c|}{0.854}          & \multicolumn{1}{c|}{0.882}          & \multicolumn{1}{c|}{0.859}          & \multicolumn{1}{c|}{14}                                         & \multicolumn{1}{c|}{0.958}          & \multicolumn{1}{c|}{0.959}          & \multicolumn{1}{c|}{0.958}          &    76  \\ \hline
ResNet-50                                                                   & pre-trained                                                                & \multicolumn{1}{c|}{0.806}          & \multicolumn{1}{c|}{0.821}          & \multicolumn{1}{c|}{0.807}          &  \multicolumn{1}{c|}{15}                                        & \multicolumn{1}{c|}{{1.000}} & \multicolumn{1}{c|}{{1.000}} & \multicolumn{1}{c|}{{1.000}} &   208      \\ \hline
ResNet-152                                                                  & pre-trained                                                                & \multicolumn{1}{c|}{0.944}          & \multicolumn{1}{c|}{0.944}          & \multicolumn{1}{c|}{0.944}          &      \multicolumn{1}{c|}{39}                                    & \multicolumn{1}{c|}{0.874}          & \multicolumn{1}{c|}{0.879}          & \multicolumn{1}{c|}{0.874}          &    521     \\ \hline
Xception                                                                    & pre-trained                                                                & \multicolumn{1}{c|}{0.938}          & \multicolumn{1}{c|}{0.944}          & \multicolumn{1}{c|}{0.938}          &    \multicolumn{1}{c|}{17}
& \multicolumn{1}{c|}{{1.000}} & \multicolumn{1}{c|}{{1.000}} & \multicolumn{1}{c|}{{1.000}} &    251     \\ \hline
ConvNext-Tiny [42]                                                              & pre-trained                                                                & \multicolumn{1}{c|}{0.938}          & \multicolumn{1}{c|}{0.938}          & \multicolumn{1}{c|}{0.938}          &    \multicolumn{1}{c|}{2214}                                      & \multicolumn{1}{c|}{{1.000}} & \multicolumn{1}{c|}{{1.000}} & \multicolumn{1}{c|}{{1.000}} &     2572    \\ \hline
SAM-Net {[}29{]}                                                            & from scratch                                                               & \multicolumn{1}{c|}{0.910}          & \multicolumn{1}{c|}{0.910}          & \multicolumn{1}{c|}{0.918}          &   \multicolumn{1}{c|}{5.2}                                       & \multicolumn{1}{c|}{0.995}          & \multicolumn{1}{c|}{0.995}          & \multicolumn{1}{c|}{0.995}          &     6.1    \\ \hline
ISA-Net$+$CBAM {[}27{]}                                                       & from scratch                                                               & \multicolumn{1}{c|}{0.676}          & \multicolumn{1}{c|}{0.666}          & \multicolumn{1}{c|}{0.676}          &      \multicolumn{1}{c|}{16}                                    & \multicolumn{1}{c|}{0.531}          & \multicolumn{1}{c|}{0.531}          & \multicolumn{1}{c|}{0.531}          &    733     \\ \hline
\begin{tabular}[c]{@{}c@{}}ViT {[}19{]}\end{tabular} & pre-trained                                                                & \multicolumn{1}{c|}{0.847}               & \multicolumn{1}{c|}{0.850}               & \multicolumn{1}{c|}{0.847}               & \multicolumn{1}{c|}{12}          & \multicolumn{1}{c|}{0.981}               & \multicolumn{1}{c|}{0.981}               & \multicolumn{1}{c|}{0.981}               &  145       \\ \hline
\begin{tabular}[c]{@{}c@{}}Pooling-ViT (PiT){[}24{]}\end{tabular} & pre-trained                                                                & \multicolumn{1}{c|}{0.847}          & \multicolumn{1}{c|}{0.858}          & \multicolumn{1}{c|}{0.848}          & \multicolumn{1}{c|}{\textbf{3.5}}          & \multicolumn{1}{c|}{0.955}          & \multicolumn{1}{c|}{0.958}          & \multicolumn{1}{c|}{0.956}          &  {3.8}       \\ \hline
\textbf{\begin{tabular}[c]{@{}c@{}}Proposed MSAD-Net\end{tabular}}  & \multicolumn{1}{l|}{\textbf{from scratch}}                                 & \multicolumn{1}{c|}{\textbf{0.986}} & \multicolumn{1}{c|}{\textbf{0.986}} & \multicolumn{1}{c|}{\textbf{0.986}} & \multicolumn{1}{c|}{{6}} & \multicolumn{1}{c|}{{1.000}} & \multicolumn{1}{c|}{{1.000}} & \multicolumn{1}{c|}{{1.000}} &     9 \\ \hline
\end{tabular}
}
\end{center}
\end{table*}

\section{Results and Analysis}
The Results and Analysis section can be summarized in four major parts: (A) Datasets and Challenges, (B) Training Specifications, (C) Performance comparisons and analysis of the state-of-the-art methods, (D) Validity checking by \textit{Explainable AI} and 5-fold cross-validation experiment. For the performance evaluation of these models, we employed ``precision" and ``F1 score" along with ``accuracy", in order to check whether the model performance is heavily influenced by class imbalance problem or not. Additionally, one more metric ``secs/epoch" is introduced in TABLE-I to indicate the average time (in secs) taken by the model per epoch during its training phase. \textbf{Hence, this metric ``secs/epoch" is analogous to the model's time complexity}. Furthermore, the graphs of all metrics vs epochs, classification reports, and confusion matrices of all these models can be found in the following GitHub link: 
\textbf{(https://github.com/Singhsshweta/Lung-Cancer-Detection)}.

\subsection{Datasets and Challenges}
We have implemented the proposed framework on two diverse CT datasets for checking validity of the proposed framework. (I) The first CT dataset is available in Kaggle: https://www.kaggle.com/datasets/kabil007/lungcancer4types-imagedataset/. This dataset encompasses four classes: adeno-carcinoma (AC), large cell carcinoma (LCC), normal, and squamous cell carcinoma (SCC). These images are split into 613 number of training, 72 validation and 315 testing samples. The number of training images in the AC, LCC, normal, and SCC classes are 195, 115, 148, and 155 respectively. Hence, there is slight class imbalance problem, additionally, the number of total training images are too less, that is 613 images. Therefore, a Deep Learning model with large number of parameters often will exhibit overfitting on this dataset. Moreover, many images from LCC have low luminance, making it a more difficult challenge. Furthermore, we have observed a higher inter-class similarity between AC and SCC class. S.Roy et al. [38] previously reported such similarity by computing correlation co-efficient between images from two different classes. All these factors make this 4-class CT dataset very much challenging. (II) The second dataset ``IQ-OTH/NCCD lung cancer" [39] was also employed in this research. This dataset was collected at the national center for cancer diseases in Iraq over a three-month period during the fall of 2019. The dataset consists of 1,190 image slices derived from 110 patient cases. This number of images in this dataset is sufficient for training heavy deep learning models. It comprises three distinct categories: normal, benign, and malignant. The distribution of training samples includes 416 images for the normal class, 120 for benign, and 561 for malignant, indicating a mild class imbalance. Additionally, the inter-class similarity [38] between any two classes was comparatively minimal in this dataset, thereby making the classification task less challenging than that of the $1^{st}$ dataset.

\subsection{Training Specifications}
All of the CNN models have been built using Keras sequential API. Tesla P100 GPU was provided by Google Colab Pro or Kaggle platform. 
	The following training specifications are followed for overall all the existing CNN models. 
	
\begin{enumerate}
		\item Both CT datasets are shuffled and then split into training, validation, and testing with a ratio of 60\%-20\%-10\%. This splitting is done in a stratified way.
		\item The The Adams-optimizer is chosen as the default optimization method for all models.
		\item All CNN and ViT models are trained using a batch size of 16.
        \item Before feeding the input images to the CNN models, all images are uniformly resized to 224$\times$224.
        \item To mitigate overfitting and maintain consistency with the proposed model, no fully connected (dense) layers are incorporated in any of the CNN models.
        \item A fixed learning rate of $1e^{-4}$ is adopted while training all the standard pre-trained CNN models. Moreover, early stopping with a patience of 5 epochs, monitored on the validation loss, is applied to prevent overfitting. 
        \item For the other trained-from-scratch models and proposed models, early stopping (es) is avoided because it causes early termination of training. Thus, all trained-from-scratch models are trained with fixed $lr$ $1e^{-4}$ and they were trained for 35 epochs.
        \item We have also incorporated adaptive learning rate (alr) in the proposed framework, in which for the first 7 epochs we maintain fixed lr $1e^{-4}$, thereafter, it will be decaying by a factor of 0.95 upto 35 epochs. This adaptive learning is part of our model novelty, and only employed in the proposed model.
        \item All these aforementioned hyper-parameters are chosen empirically. No automated grid search method is employed here.
        \item Categorical Cross Entropy (CCE) is employed as a loss function for all the benchmark models. 
        \item No pre-processing method, like data-augmentation, is deployed in any of the experiments. 	
\end{enumerate}
\subsection{Performance comparisons of the State-of-the-art methods} 
The proposed framework, ``MSAD-Net", is evaluated against several state-of-the-art approaches as presented in TABLE-I. Specifically, its performance is compared with standard pre-trained CNN architectures, including DenseNet-121 [40], Inception-V3 [41], MobileNet-V2 [36], ResNet-50 [33], ResNet-152, Xception, and latest CNN model ``ConvNext Tiny"[42]. In addition, the framework is also compared with the recent transformer-based models such as the Vision Transformer (ViT) [19], and Pooling-based ViT (PiT) [24]. Furthermore, the performance of two recently proposed CNN architectures ``ISA-Net" [27] and ``SAM-Net" [29], trained from scratch, is reported in TABLE-I. Out of these models, ViT [19], and ``ISA-Net" [27] were specifically designed and proposed for lung cancer classification from CT datasets. 

As seen in TABLE-I, ISA-Net performs poorly on both CT datasets, likely due to its heavy architecture (comprising 14.1 million parameters) and the fact that it was trained from scratch, which makes the model susceptible to overfitting. In contrast, SAM-Net, another trained from scratch model, demonstrates higher efficacy across both CT datasets. This can be attributed to its lightweight design.
Although SAM-Net had decent performance on challenging 4-class CT dataset (around 91\% accuracy), it could not surpass the efficacy of the proposed model. TABLE-I demonstrates that the proposed framework ``MSAD-Net" not only outperformed these two trained-from-scratch models, but also, demonstrated superior performance compared to all other pre-trained models and recent trend Vision Transformer models.

\begin{table*}[t]
		\begin{center}		\caption{Comparison of class-wise classification reports (on testing), of numerous recent trend models with the proposed MSAD-Net. Implementation is done on \textbf{``4-class CT dataset"}}
		\resizebox{1.01\columnwidth}{!}{
\begin{tabular}{|c|ccc|ccc|ccc|ccc|}
\hline
\multirow{2}{*}{Classes}                               & \multicolumn{3}{c|}{Vision Transformer (ViT)}                                            & \multicolumn{3}{c|}{Pooling based ViT (PiT)}                          & \multicolumn{3}{c|}{SAM-Net}                                     & \multicolumn{3}{c|}{\textbf{MSAD-Net}}                                               \\ \cline{2-13} 
                                                       & \multicolumn{1}{c|}{Precision} & \multicolumn{1}{c|}{Recall} & F1-score & \multicolumn{1}{c|}{Precision} & \multicolumn{1}{c|}{Recall} & F1score & \multicolumn{1}{c|}{Precision} & \multicolumn{1}{c|}{Recall} & F1-score & \multicolumn{1}{c|}{Precision}      & \multicolumn{1}{l|}{Recall}         & F1-score       \\ \hline
AC                                                     & \multicolumn{1}{c|}{0.789}     & \multicolumn{1}{c|}{0.875}  & 0.830    & \multicolumn{1}{c|}{0.845}     & \multicolumn{1}{c|}{0.816}  & 0.830   & \multicolumn{1}{c|}{0.980}     & \multicolumn{1}{c|}{0.820}  & 0.890    & \multicolumn{1}{c|}{0.980}          & \multicolumn{1}{c|}{0.980}          & 0.980          \\ \hline
LCC                                                  & \multicolumn{1}{c|}{0.826}     & \multicolumn{1}{c|}{0.745}  & 0.783    & \multicolumn{1}{c|}{0.700}     & \multicolumn{1}{c|}{0.960}  & 0.809   & \multicolumn{1}{c|}{0.900}     & \multicolumn{1}{c|}{0.960}  & 0.930    & \multicolumn{1}{c|}{0.960}          & \multicolumn{1}{c|}{0.960}          & 0.960          \\ \hline
Normal                                                     & \multicolumn{1}{c|}{1.000}     & \multicolumn{1}{c|}{0.981}  & 0.990    & \multicolumn{1}{c|}{1.000}     & \multicolumn{1}{c|}{0.981}  & 0.990   & \multicolumn{1}{c|}{0.940}     & \multicolumn{1}{c|}{0.970}  & 0.950    & \multicolumn{1}{c|}{1.000}          & \multicolumn{1}{c|}{1.000}          & 1.000          \\ \hline
SCC                                                 & \multicolumn{1}{c|}{0.855}     & \multicolumn{1}{c|}{0.789}  & 0.820   & \multicolumn{1}{c|}{0.881}     & \multicolumn{1}{c|}{0.744}  & 0.807   & \multicolumn{1}{c|}{0.840}     & \multicolumn{1}{c|}{0.950}  & 0.890    & \multicolumn{1}{c|}{1.000}          & \multicolumn{1}{c|}{1.000}          & 1.000         \\ \hline
Macro-Avg                                              & \multicolumn{1}{c|}{0.867}     & \multicolumn{1}{c|}{0.847}  & 0.856   & \multicolumn{1}{c|}{0.856}     & \multicolumn{1}{c|}{0.875}  & 0.859   & \multicolumn{1}{c|}{0.910}     & \multicolumn{1}{c|}{0.920}  & 0.910    & \multicolumn{1}{c|}{\textbf{0.990}} & \multicolumn{1}{c|}{\textbf{0.990}} & \textbf{0.990} \\ \hline
\begin{tabular}[c]{@{}c@{}}Weight-Avg\end{tabular} & \multicolumn{1}{c|}{0.850}     & \multicolumn{1}{c|}{0.847}  & 0.847    & \multicolumn{1}{c|}{0.858}     & \multicolumn{1}{c|}{0.847}  & 0.848   & \multicolumn{1}{c|}{0.920}     & \multicolumn{1}{c|}{0.910}  & 0.910    & \multicolumn{1}{c|}{\textbf{0.990}} & \multicolumn{1}{c|}{\textbf{0.990}} & \textbf{0.990} \\ \hline
\end{tabular}
}
\end{center}
\end{table*}
\begin{figure*}[h]
		\centering
        \includegraphics[width=18.2cm, height=4.7cm]{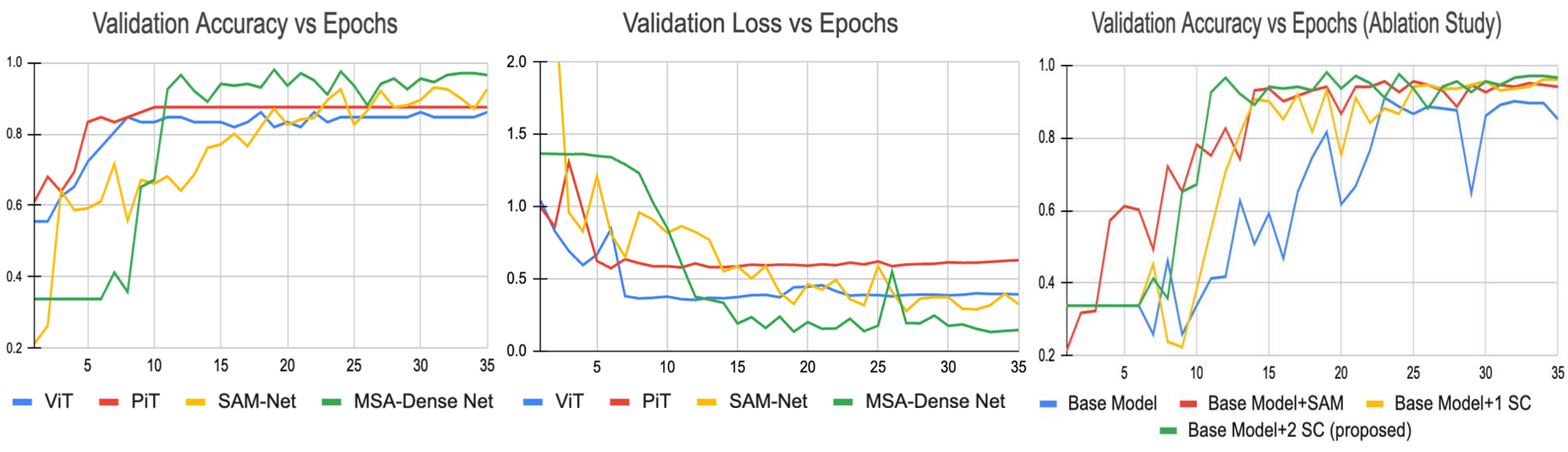}
		\caption{Graph comparison of proposed model with the recent trend models on 4-class CT dataset. From left to right: graph of (I) Validation accuracy vs epochs, (II) Validation loss vs epochs, (III) Validation accuracy vs epochs (ablation studies). \textbf{For better visualization, zooming is preferable}.}
	\end{figure*}
    \begin{table*}[tb]
\begin{center}
\caption{Ablation Studies of the proposed ``MSAD-Net'' on 4-class CT dataset}
\label{tab:my-table}
\resizebox{1.00\columnwidth}{!}
{
\begin{tabular}{|c|c|c|c|c|c|c|c|}
\hline
\begin{tabular}[c]{@{}c@{}}Methodology/ \\ Performance\end{tabular} & \begin{tabular}[c]{@{}c@{}}Base Model\\ with no SC\end{tabular} & \begin{tabular}[c]{@{}c@{}}Base Model\\$+$ 1SC\end{tabular} & \begin{tabular}[c]{@{}c@{}}Base Model\\ $+$SAM by {[}29{]}\end{tabular} & \begin{tabular}[c]{@{}c@{}}Base Model$+$2SC \\ (with 64 no of \\ 5x5 conv in last SC)\end{tabular} & \begin{tabular}[c]{@{}c@{}}Base Model$+$2SC\\ (with 32 no of \\ DilConv in last SC)\end{tabular} & \begin{tabular}[c]{@{}c@{}}Base Model$+$2SC\\ (with 64 no of \\ DilConv in last SC)\end{tabular} & \begin{tabular}[c]{@{}c@{}}Base Model$+$2SC \\ (with 96 no of \\ DilConv in last SC)\end{tabular} \\ \hline
Accuracy                                                     & 0.876                                                           & 0.938                                                      & 0.924                                                                  & 0.958                                                                                            &   0.910                                                                                             & 0.972                                                                                           & \textbf{0.986}                                                                                  \\ \hline
Precision                                                    & 0.876                                                           & 0.944                                                      & 0.939                                                                  & 0.958                                                                                            &   0.916                                                                                             & 0.972                                                                                           & \textbf{0.986}                                                                                  \\ \hline
Recall                                                       & 0.876                                                           & 0.937                                                      & 0.924                                                                  & 0.958                                                                                            &  0.903                                                                                              & 0.972                                                                                           & \textbf{0.986}                                                                                  \\ \hline
F1-score                                                     & 0.876                                                           & 0.938                                                      & 0.931                                                                  & 0.958                                                                                            &  0.910                                                                                              & 0.972                                                                                           & \textbf{0.986}                                                                                  \\ \hline
AUC                                                          & 0.982                                                           & 0.991                                                      & 0.990                                                                  & 0.993                                                                                            &    0.990                                                                                            & \textbf{0.999}                                                                                  & \textbf{0.999}                                                                                  \\ \hline
\end{tabular}
}
\end{center}
\end{table*}

 As previously discussed in Section III-A, the 3-class CT dataset is relatively less challenging, thus, majority of the pre-trained CNN models and ViT models exhibit strong performance on this CT dataset, demonstrated in TABLE-I. In contrast, the 4-class CT dataset presents a greater level of complexity; therefore, certain pre-trained CNN models such as DenseNet, MobileNet-V2, and ResNet-50 suffered from poor testing efficacy on this dataset. Overall, Inception-V3, ResNet-152, Xception, and ConvNeXt-T demonstrated relatively better performance on the 4-class dataset, but none of them could exceed the accuracy or F1-score of 94\%. On the contrary, our proposed MSAD-Net (trained from scratch) has achieved the highest testing accuracy (98.6\%), precision (98.6\%), and F1-score (98.6\%), which is so far the best performance on this challenging 4-class CT dataset. Moreover, it generalized well on the 3-class dataset and achieved 100\% testing accuracy just like ConvNext-T, Xception and ResNet-50 model. This was possible because our proposed ``MSAD-Net" attributed very less number of trainable parameters (1.1 million only) and due to the incorporation of novel SAM attention block, which enables the model to effectively capture and preserve critical global features of lung cancer patterns. As illustrated in TABLE-I, the Vision transformer could not exceed 85\% testing accuracy, due to slight overfitting occurring likely for to its heavy-weight architecture. Pooling-based ViT (PiT) had  comparatively a lightweight model than ViT, however, it demonstrated a very similar efficacy in TABLE-I. This indicates that the vision transformer models have not resolved the challenges of small ``4-class CT dataset". It is important to note that for a model which is entirely trained from scratch, consistently exceeding the performance of all pre-trained models is a challenging task. Nevertheless, our proposed model has successfully achieved this milestone, with a very less computational complexity. It can be observed from TABLE-I that the time-complexity (secs/sp) by our proposed model is significantly lesser than that of other models. Only SAM-Net and PiT showed similar time complexity as our model. 

\begin{figure*}[h]
		\centering
		\includegraphics[width=17.0cm,height=6.0 cm]{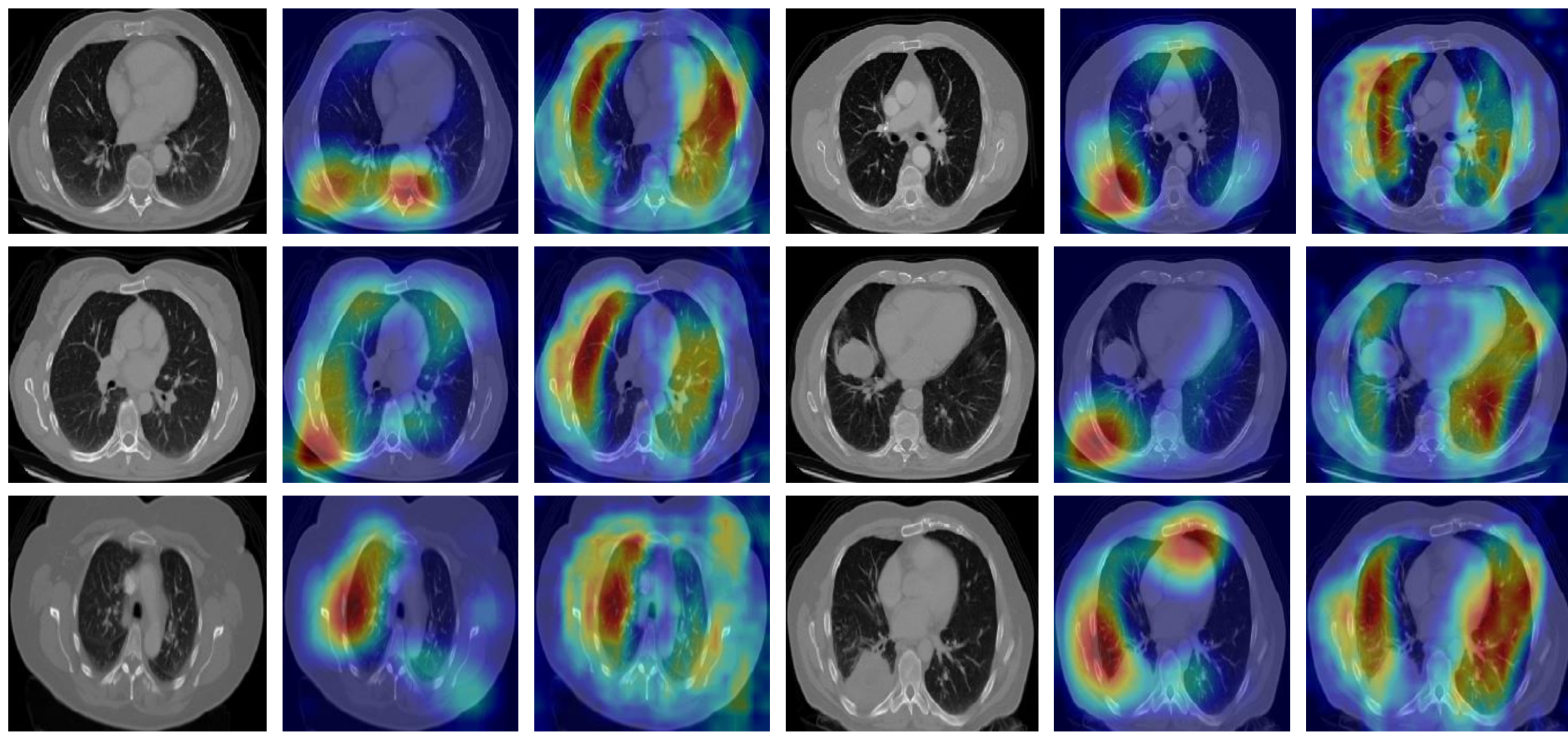}
		\caption{\textbf{Validity checking of MSAD-Net by Explainable AI:} First clumn images represent Original Lung cancer CT images, $2^{nd}$ column represents Gradcam heat map by the proposed MSAD-Net model without SAM, (c) $3^{rd}$ column represents Gradcam heat map by the proposed MSAD-Net model with SAM.}
	\end{figure*}

Moreover, a detailed class-wise classification report is provided in TABLE-II (for only 4-class CT dataset), showcasing the class-wise performance of MSAD-Net in comparison with several recently proposed models, including recent trend ViT, PiT, and SAM-Net. From this kind of class-wise representation in TABLE-II, it is now easier to visualize whether any model is effected by class-imbalance or not. Moreover, TABLE-II reveals that our proposed framework, ``MSAD-Net" not only outperforms the overall efficacy of the other three existing models, but also achieves superior class-wise results, along with higher macro and weighted averages of precision, recall, and F1 score individually, thereby providing more consistent and reliable results on challenging 4-class CT dataset. 

Furthermore, the validation graphs of ViT, PiT and SAM-Net are compared with the proposed MSAD-Net, as illustrated in Fig.4. 
This is to clarify that we deliberately excluded the performance of latest models such as ISA-Net and ConvNext-T in the graph of Fig.4, to avoid considerable fluctuations (or incompatibility) in the graph. From these plots in Fig.4, it is evident that both ViT and PiT models converged to higher validation accuracy or lower validation loss earlier than the other two models. However, their performance plateaued after just a few epochs (around 5–10), likely due to overfitting. This ``overfitting" condition can be further verified from the training graphs given in the anonymous GitHub repository (mentioned in the abstract). As illustrated in Fig. 4, SAM-Net delivers a decent and competitive performance compared to the proposed framework; however, its validation graph performance (in terms of accuracy and loss) slightly falls just short of surpassing the proposed 'MSAD-Net' (represented in green). This detailed class-wise performance representation, in TABLE-II along with graphs in Fig.4, has strengthened the credibility and validity of our experimental findings.

Moreover, an ablation study of ``MSAD-Net", conducted on 4-class CT dataset, is presented in TABLE-III. This ablation study, supported by the graphs depicted in the $3^{rd}$ column of Fig.4 (i.e., ablation study graph), provides additional justification for the design choice of the proposed ``MSAD-Net" architecture. For instance, from the TABLE-III and the Validation graph (ablation study), it is evident that after adding the $1^{st}$ Skip Connection (SC), the performance of the base model is improved significantly (by approximately 6\%), resulting in a more stable (validation) graph than the previous. Furthermore, the inclusion of the Spatial Attention Module (SAM) as a $2^{nd}$ Skip Connection (SC), further boosted the model’s performance considerably (from 93.8\% to 98.6\%). These are notable boosting performance. In particular, we have experimented with the several combinations of numbers of DilConv filters which were mainly responsible for multiscale attention. Empirically, we found that while utilizing 64 and 96 number of DilCov layers, the proposed model has achieved the best efficacy in the range of $97.2\% - 98.6\%$, hence, they justified the particular design choice for the proposed ``MSAD-Net".

\subsection{Validity checking by Explainable AI and 5-fold cross validation}
To validate our proposed theory, we have implemented proposed ``MSAD-Net" with the help of \textit{Explainable AI}.
A Grad-CAM heat map visualization [43] is presented in Fig.5. Here, in the $1^{st}$ column, the CT images are taken from both 4-class and 3-class CT datasets. From the $2^{nd}$ column of Fig.5, it is evident that the proposed base model (without SAM) could not consistently give attention directly to the lung modules or region of interest. In contrast, in the $3^{rd}$ column of Fig.5, it is apparent that proposed base model along with SAM, focuses more intensely on the lung regions and sometimes it directly focus on the suspicious portions of those lung regions. This experiment validates that integrating the proposed SAM block enables the base CNN model to focus directly to the lung segments, thus significantly increasing the model’s efficacy. 

\begin{table}[t]
		\begin{center}		\caption{Testing results of proposed MSAD-Net, for 5-fold cross validation on `4-class CT Dataset' (\textbf{Weighted Average is displayed})}
            \vspace{0.2cm}
		\resizebox{1.01\columnwidth}{!}{
		\begin{tabular}{|c|c|c|c|c|c|}
			\hline
			{folds} & {Accuracy} & {Precision} & {Recall} & {F1-score} & {AUC}\\
                \hline
                \hline
			fold1  & 0.960 & 0.960 & 0.960 & 0.960 & {0.998}\\
			\hline
			fold2 & 0.975 & 0.975 & 0.975 & 0.975 & 0.999\\
			\hline
   fold3  & {0.965} & {0.965} & {0.965}& {0.965} & 0.988\\
			\hline
			fold4 & {0.955} & {0.955} & {0.955} & {0.955} & 0.990\\
   \hline
   fold5  & 0.970 & 0.970 & 0.970 & 0.970 & 0.986\\
			\hline
   \textbf{Mean $\pm$ Std} & \textbf{0.964}             & \textbf{0.964}             & \textbf{0.964}             & \textbf{0.964} & \textbf{0.992}\\
  \textbf{Deviation} & \textbf{$\pm$ 0.015} & \textbf{$\pm$ 0.015} & \textbf{$\pm$ 0.015} & \textbf{$\pm$ 0.015} & \textbf{$\pm$ 0.006} \\
   \hline
		\end{tabular}
  }
	\end{center}
\end{table}

For validity of our proposed method, we have also conducted a stratified 5-fold cross-validation experiment on the ``4-class CT Dataset" for proposed ``MSAD-Net". We effectively created the equivalent of 5 different datasets (we call them fold1-to-fold5). Here each dataset has a distinct testing set, having different statistics compared to the same of the other 4 datasets. Hence, this environment is even more challenging than the ``4-class CT dataset". The results of this 5-fold cross-validation, with mean and standard deviation values, are presented in TABLE-IV. It is evident from TABLE-IV the proposed framework has attained a mean of 96.4\% accuracy, precision, recall and F1-score with standard deviation ($<=1.5\%$). This reveals that the proposed framework consistently performed over all 5 diverse folds of CT datasets, with very less deviation. Hence, this experiment proves the validity or reliability of the proposed model in this research. 

\section{Conclusion and Future Works}
A Multiscale and Spatial Attention-based Dense Network (MSAD-Net) was proposed for the efficient classification of several types of lung cancers from CT images. A dense module, incorporated in the proposed model, used DWSC layers and 1$\times$1 layers, to significantly reduce the number of trainable parameters. Additionally, a novel spatial attention module (SAM) was leveraged on top of this base model. Unlike traditional SAM which employed a single 7×7 convolutional layer, the proposed SAM incorporated multiple numbers of dilated convolutional filters with a dilation rate of 2. 
Extensive results showed that the proposed SAM module boosted the efficacy of the base CNN model substantially with less computation. For validity of our proposed model, ``MSAD-Net" was also implemented by \textit{``Explainable AI"}. This experiment substantiated our theory that SAM block enabled the proposed base model to focus more on the suspicious lung module. Furthermore, a stratified 5-fold cross-validation experiment was conducted for the proposed ``MSAD-Net". These experimental evaluations demonstrated that the proposed model, ``MSAD-Net", not only outperformed the recent trend models by significant margin but also exhibited strong generalization across diverse 5-fold datasets.

Although from CT images deep learning models can predict several types of lung cancers very efficiently, collecting CT scans images are are typically expensive. Moreover, in comparison to CT scan images, histopathology images hold greater promise for detecting small lung cancer at much earlier stages. Thus, in the future, we aim to extend this project to ``early detection of lung cancers using complex histopathology datasets".

%



\ifCLASSOPTIONcaptionsoff
  \newpage
\fi

\end{document}